\documentclass[conference]{IEEEtran}
\IEEEoverridecommandlockouts
\usepackage{cite}
\usepackage{amsmath,amssymb,amsfonts}
\usepackage{graphicx}
\usepackage{textcomp}
\usepackage{xcolor}
\usepackage{hyperref}
\usepackage{url}
\usepackage{booktabs}
\usepackage{cite}
\usepackage{pgfplots}
\pgfplotsset{compat=1.18}
\usetikzlibrary{patterns,patterns.meta}
\usepackage{amsmath,amssymb,amsfonts}
\usepackage{graphicx}
\usepackage[ruled,vlined]{algorithm2e}
\usepackage{makecell}
\usepackage{tikz}
\usepackage{pgfplots}
\pgfplotsset{compat=1.18}
\def\BibTeX{{\rm B\kern-.05em{\sc i\kern-.025em b}\kern-.08em
    T\kern-.1667em\lower.7ex\hbox{E}\kern-.125emX}}

\makeatletter
\def\ps@IEEEtitlepagestyle{%
  \def\@oddhead{\hfill \footnotesize Accepted for publication in the \textit{2026 IEEE International Conference on Big Data (Big Data)}, Phoenix, AZ, USA, December 14-17, 2026.}%
  \def\@oddfoot{\mycopyrightnotice}%
  \def\@evenhead{}%
  \def\@evenfoot{}%
}
\def\mycopyrightnotice{%
  {\begin{minipage}{\textwidth}\centering\scriptsize
  © 2026 IEEE. Personal use of this material is permitted. Permission from IEEE must be obtained for all other uses, in any current or future media, including reprinting/republishing this material for advertising or promotional purposes, creating new collective works, for resale or redistribution to servers or lists, or reuse of any copyrighted component of this work in other works.
  \end{minipage}}%
}
\makeatother
\begin{document}

%\title{SyntheIce: Robust Synthetic Radargram Benchmark Dataset and Slope-aware Fragmented Englacial Layer Reconstruction}

%\title{SyntheIce: Synthetic Radargram Annotation Benchmark and Slope-aware Fragmented Englacial Layer Reconstruction}

\title{SyntheIce: A Synthetic Radargram Annotation Benchmark for Englacial Layer Reconstruction}

%\title{SyntheIce: Synthetic Radargrams for Englacial Layer Reconstruction}

\author{
\IEEEauthorblockN{Muhammad Behroze Hassan\IEEEauthorrefmark{1} \IEEEauthorrefmark{2}, Bayu Adhi Tama\IEEEauthorrefmark{2}, Sanjay Purushotham\IEEEauthorrefmark{1}  \IEEEauthorrefmark{2}, Vandana P. Janeja\IEEEauthorrefmark{1}  \IEEEauthorrefmark{2}}
\IEEEauthorblockA{\IEEEauthorrefmark{1}Department of Information Systems, University of Maryland Baltimore County (UMBC), Maryland, USA}
\IEEEauthorblockA{\IEEEauthorrefmark{2}iHARP, University of Maryland Baltimore County (UMBC), Maryland, USA\\ 
\{mhassan4; bayu; psanjay, vjaneja\}@umbc.edu}
}
\maketitle

%%
%% By default, the full list of authors will be used in the page
%% headers. Often, this list is too long, and will overlap
%% other information printed in the page headers. This command allows
%% the author to define a more concise list
%% of authors' names for this purpose.

%%
%% The abstract is a short summary of the work to be presented in the
%% article.
\begin{abstract}

Englacial layers in radio-echo-sounding radargrams (subsurface ice images)
encode the past climate and flow history of ice sheets and glaciers, but
extracting them at scale is bottlenecked by time-consuming manual annotation and
by semi-automated detectors that return broken, disconnected fragments rather
than continuous horizons. We address two coupled problems in layer annotation products: repairing partial annotations by connecting their fragments into coherent full-layer reconstructions across an entire profile, and validating layer continuity across the boundaries of adjacent tiles. We introduce a slope-aware greedy fragment-linking algorithm that reconstructs full-width layers from partial
detections, assigning each fragment to its parent layer through row-similarity
gating and an exponentially weighted slope predictor (EWMA), and attaching a
content-based hash ID to each reconstruction for stable cross-radargram
identification. For development and rigorous quantitative evaluation, we generate
a synthetic dataset of 600 radargrams spanning multiple complexity levels,
including noise close to realistic conditions, 150 are wide radargrams cropped
into four adjacent tiles. All come with exact ground-truth annotations, giving
full control for validation. We test this dataset with XCheck, a
consistency-based validation framework: whereas XCheck previously operated on
only two consecutive images, we apply it across four consecutive tiles,
enforcing continuity at every boundary, which also indicates that the synthetic
data is realistic enough to drive real-world algorithms. Our major finding
generalizes beyond ice: when linear features in spatial data are broken into
fragments but keep a consistent local geometry, linking them by predicted
trajectory is more reliable than by appearance. Evaluated against expert-labeled
real radargrams, the method achieves high precision, recall, F1, and mean IoU,
establishing transfer of our findings to a real-world dataset.
\end{abstract}


%%
%% The code below is generated by the tool at http://dl.acm.org/ccs.cfm.
%% Please copy and paste the code instead of the example below.
%%

%% A "teaser" image appears between the author and affiliation
%% information and the body of the document, and typically spans the
%% page.
%\begin{teaserfigure}
%\includegraphics[width=\textwidth]{sampleteaser}
%  \caption{Seattle Mariners at Spring Training, 2010.}
%  \Description{Enjoying the baseball game from the third-base
%  seats. Ichiro Suzuki preparing to bat.}
%  \label{fig:teaser}
%\end{teaserfigure}

%%
%% This command processes the author and affiliation and title
%% information and builds the first part of the formatted document.
\maketitle

\section{Introduction}
\label{sec:introduction}

The internal structure of the Greenland and Antarctic ice sheets is one of the most information-rich archives available to glaciology. The deep buried englacial layers, imaged as bright reflectors in ice-penetrating radar, are
widely interpreted as isochrones: surfaces of equal age that trace past snow layers, recording ice flow, accumulation, and deformation over time~\cite{macgregor2015},~\cite{holschuh2017,moqadam2025review}. Because the geometry of these layers encodes boundary conditions that are otherwise difficult to observe, their large-scale tracing has become central to dating ice cores,
calibrating ice-flow models, and reconstructing past ice sheet dynamics~\cite{holschuh2017,moqadam2025review}. The primary instrument for acquiring this signal is airborne radio-echo sounding (RES), which produces two-dimensional radargrams in which the vertical axis is two-way travel time (depth) and the horizontal axis is distance along the flight track.

Despite decades of data collection, converting raw radargrams into usable layer geometry remains a bottleneck. Expert manual tracing is accurate but extremely
labor-intensive, and it scales poorly to the millions of profile kilometres now available~\cite{macgregor2015,moqadam2025review}. This has motivated a long line of semi-automated tracing methods, progressing from image-processing and edge-detection approaches to recent deep-learning segmentation pipelines~\cite{moqadam2025review},~\cite{tama2025mlt,moqadam2025deep}. A recurring limitation, however, is that automated detectors rarely return clean, full-width layers: steeply dipping reflectors are attenuated or lost during conventional synthetic-aperture-radar (SAR) focusing~\cite{castelletti2019,arenas2025}, and noise, signal fade with depth, and crossing structures cause detectors to output fragmented skeletons, many short disconnected pieces of what is physically one continuous layer. Reassembling these fragments into coherent layers is itself an unsolved spatial problem.

Two gaps make this difficult to study and to solve. First, there is no controlled, benchmarked ground truth at the scale of radargram annotations: existing synthetic radar modeling is one-dimensional and tied to a few ice-core sites, converting measured profiles into synthetic traces to explain the physical origin of reflections rather than producing full two-dimensional radargrams with known per-layer geometry~\cite{mojtabavi2022}. As a result, automated methods are evaluated against expert annotations that are sparse because manual tracing is so time consuming. Second, fragment-to-layer reconstruction needs a cue that survives gaps. Englacial layer slope (dip) is a strong, physically meaningful continuity signal~\cite{holschuh2017,castelletti2019,arenas2025}, yet most tracing pipelines do not use it to link fragments across the radar image, and they rarely enforce consistency across adjacent radargram tiles even though annotation agreement is a known quality concern~\cite{tama2024annotation,jebeli2023tssa}.
 
In this paper we close both gaps with an end-to-end framework that runs from controlled synthetic generation to validation based on expert annotations. We make the following contributions:

\begin{itemize}
  \item \textit{SyntheIce}: a controlled generator that produces synthetic englacial-layer radargram annotations with exact, per-layer ground truth, so methods can be scored against known geometry instead of noisy human labels. In addition, we examine generative AI radargram synthesis as a reference point for SyntheIce, to see how such tools render layered structure.
  \item \textit{TraceLink}: a slope-aware greedy algorithm that links fragmented detections back into full-width layers, using row-similarity gating together with an exponentially weighted moving average (EWMA) slope predictor to assign each fragment to its parent layer.
  \item \textit{A consistency-based preparation step} that crops 150 wide radargrams into four adjacent tiles each creating 600 radargrams and applies XCheck ~\cite{xcheck} to measure annotation agreement across neighbouring tiles.
  \item \textit{A controlled real-data validation}: we take expert layers spanning at least 70\% of a radargram, inject randomized gaps to mimic detector fragmentation, reconstruct them with TraceLink, and score against the original layer as ground truth; showing how well the method carries over from synthetic data to real annotations. 
\end{itemize}

The SyntheIce annotation dataset is available at:
\url{https://github.com/iharp-institute/SyntheIce-Dataset}.
The code for reproducing all experiments is available at:
\url{https://github.com/iharp-institute/SyntheIce-Englacial-Layer-Reconstruction}.

\section{Related Work}
\label{sec:related}
 
\subsection{Existing Methods and Techniques}
\label{sec:related-methods}
 
\textit{Synthetic radargram and englacial-layer data.}
Most synthetic
englacial stratigraphy comes from one-dimensional forward modeling at ice-core sites. Eisen et al.~\cite{eisen2004} convert profile measurements into synthetic radar traces to date reflectors, and Mojtabavi et al.~\cite{mojtabavi2022} apply the same idea at NGRIP, NEEM, and EGRIP on the Greenland ice sheet to explain why internal reflections appear where they do.
These models work at the level of single traces tied to a borehole, not full radargram images. A second line of work builds stratigraphy from ice-flow models: Born and Robinson~\cite{born2021} simulate Greenland isochrones with an advection scheme, later formalized as the ELSA layer-tracing model~\cite{rieckh2024}.
Their output is modeled layer elevations for comparison against data, not radar images with labeled layers. Closer to image space, Elsworth et al.~\cite{elsworth2020} produce two-dimensional synthetic radargrams to study how ice flow deforms layers, and Rahnemoonfar et al.~\cite{rahnemoonfar2019} use a generative adversarial network (GAN) to synthesize radar-sounder images for
data augmentation. The GAN images, however, carry only coarse layer masks, and the authors report that they cannot train a tracing network on their own.
 
\textit{Automated tracing and slope estimation.}
On real data, the dominant task has been tracing internal reflecting horizons (IRHs). MacGregor et al.~\cite{macgregor2015} built the reference dated radio-stratigraphy of Greenland using semi-automated peak following guided by predicted reflection slope. Image-processing trackers detect layers through ridge or peak detection and seed propagation: Sime et al.~\cite{sime2011} derive dip from layer objects without requiring continuous reflectors, and Xiong et al.~\cite{xiong2018}. (ARESELP) trace
layers from MCoRDS data near NEEM in northwest Greenland. Probabilistic and learning-based methods cast tracing as inference, from graphical models and hidden Markov models~\cite{crandall2012,carrer2017} to recent deep networks
that segment IRHs directly~\cite{varshney2021,moqadam2025deep}. Slope has also been studied on its own: Holschuh et al.~\cite{holschuh2017} show that a Radon-transform slope estimate stays reliable even where reflectors are noisy and broken, and Castelletti et al.~\cite{castelletti2019} recover
steeply dipping layers through layer-optimized SAR focusing. The MorphoLayerTrace~\cite{tama2025mlt} improves seed selection to reduce discontinuities, and XCheck~\cite{xcheck} scores annotation consistency across intersecting flight lines.

\subsection{Limitations of Current Methods and Positioning of Our Work}
\label{sec:related-gaps}
 
Two gaps follow from this study of work. First, there is no controlled synthetic dataset of two-dimensional radargrams internal layers with exact, per-layer ground truth. The forward models are one-dimensional and tied to boreholes~\cite{eisen2004,mojtabavi2022}, the ice-flow models output layer elevations rather than radar images~\cite{born2021,rieckh2024}, and the GAN
images lack precise per-layer labels~\cite{rahnemoonfar2019}. As a result, automated methods are usually scored against expert annotations whose own gaps and inconsistencies are baked into the reference. Second, tracing methods detect or follow layers but do not put broken pieces (fragments) back together. Slope is known to survive where continuous tracing fails~\cite{holschuh2017,castelletti2019}, yet it is used to estimate dip, not to decide which fragment belongs to which layer. Cross-tile agreement is rarely enforced, even though annotation consistency is a known weak point~\cite{xcheck,tama2024annotation}. Benchmarks make the same point: in a side-by-side comparison of automated trackers, Delf et al.~\cite{delf2020} recommend that future methods focus on linking picked stratigraphy across signal disruptions, and the review by Moqadam and Eisen~\cite{moqadam2025review} concludes that automated mapping still struggles with folds and discontinuities. To the best of our knowledge, no published method performs slope-aware fragment-to-layer reconstruction in
radargrams.

Our framework targets both gaps directly. SyntheIce supplies controlled two-dimensional radargrams internal structure with exact per-layer annotations, so a method can be measured against known geometry instead of noisy labels. TraceLink uses slope as the linking cue that prior work left unused: a row-similarity gate together with an exponentially weighted moving average (EWMA) slope predictor assigns each fragment to its parent layer and rebuilds full-width layers from broken detections. For the wide radargrams XCheck~\cite{xcheck} adds the consecutive radargrams consistency check that tracing pipelines usually skip. Table~\ref{tab:dataset-comparison} positions the data produced by SyntheIce against the real Greenland CReSiS dataset~\cite{paden2014mcords}, benchmarks used in layer studies available through the CReSIS data
portal~\cite{cresis2024rds}, showing that the generated categories capture key characteristics observed in real snow-radar and depth-sounder data, including variation in layer count, fragmentation, terrain tilt, and depth, while adding controlled stress cases that existing datasets do not provide
\begin{figure}[ht!]
  \centering
  \begin{minipage}{0.48\linewidth}
    \centering
    \includegraphics[width=\linewidth]{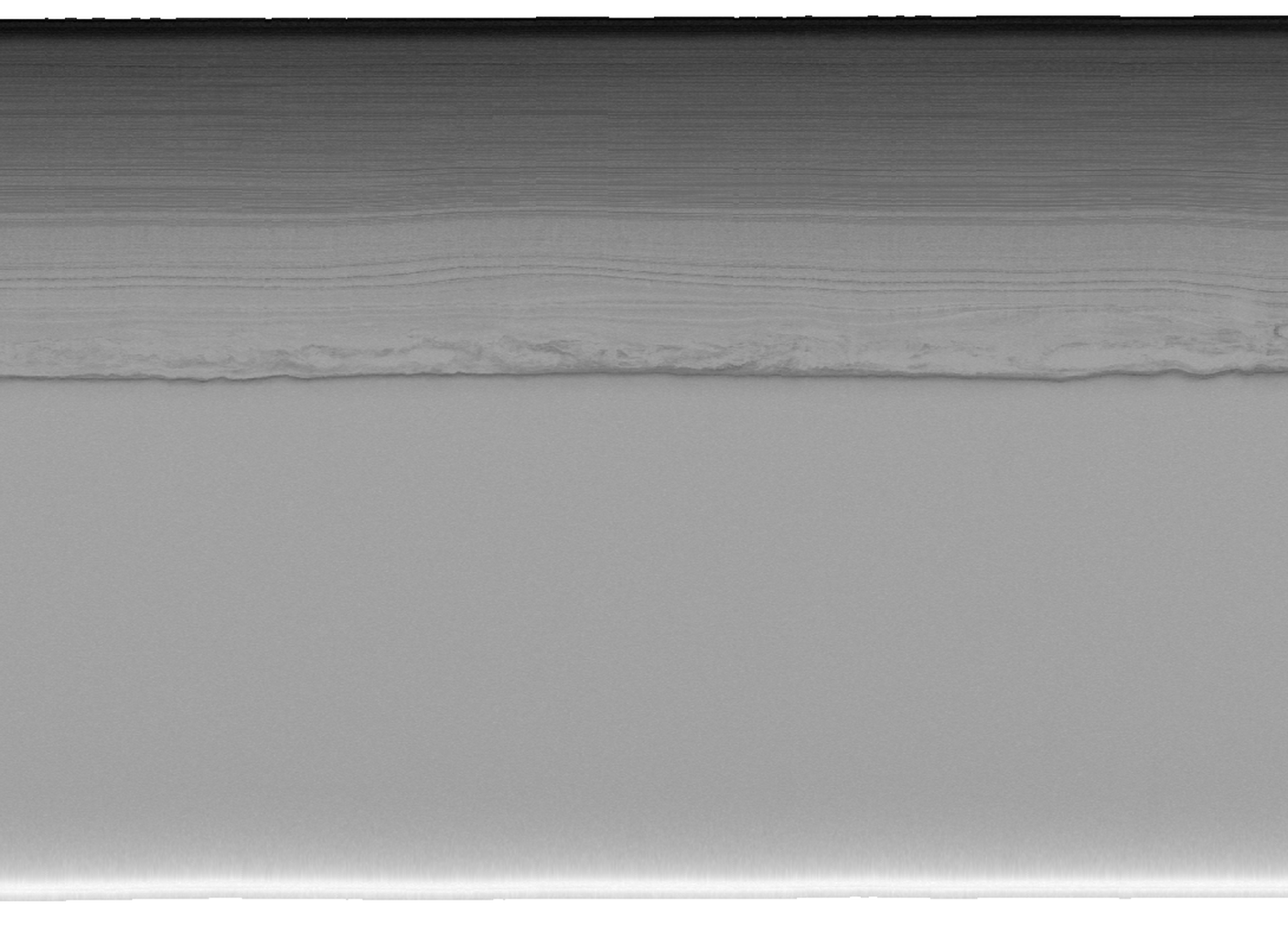}
    \\[2pt]{\footnotesize (a) input radargram}
  \end{minipage}\hfill
  \begin{minipage}{0.48\linewidth}
    \centering
    \includegraphics[width=\linewidth]{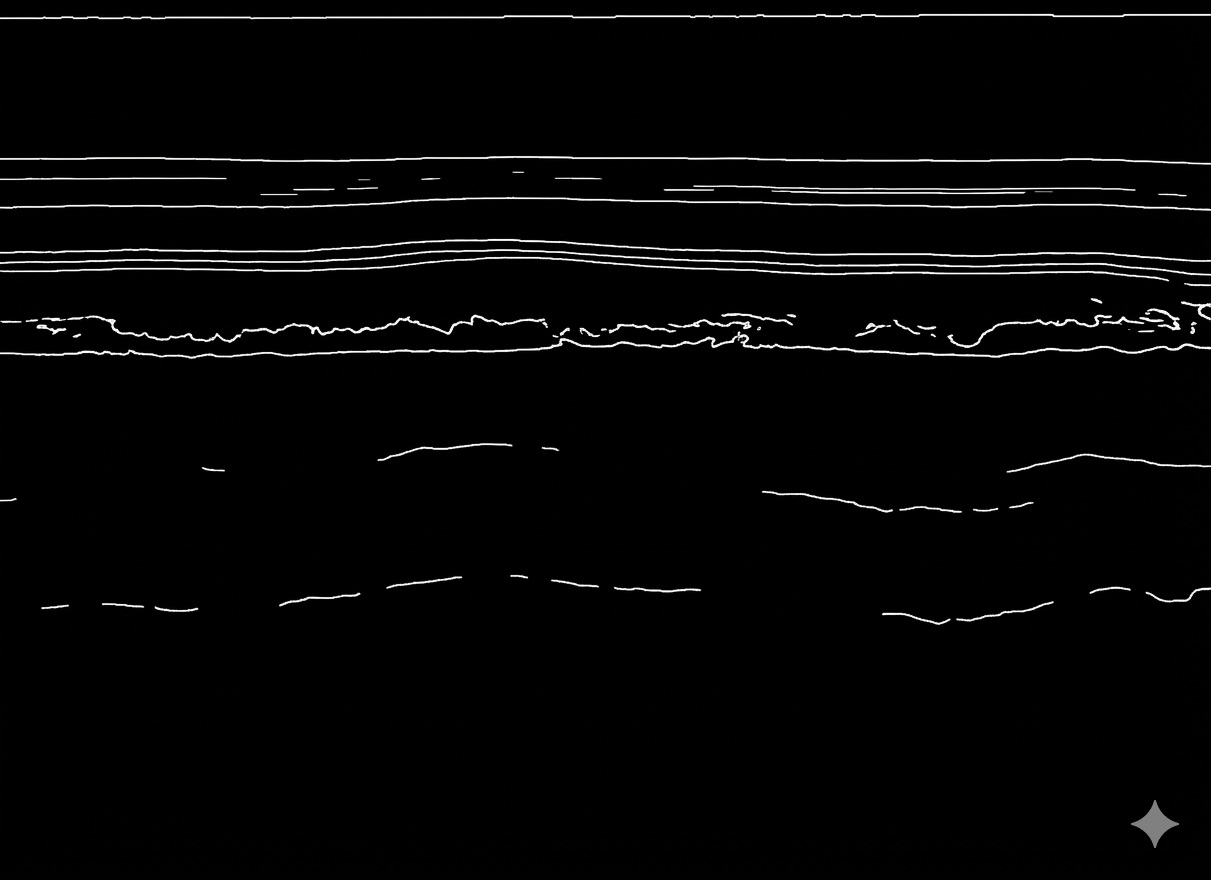}
    \\[2pt]{\footnotesize (b) generated mask}
  \end{minipage}
  \caption{A generative image model fails to annotate englacial layers.
   Given input radargram (Data\_20120514\_01\_015) (a) and expert example pairs, the model returns
  the mask in (b), which hallucinates layers that are not present and misses
  real ones. Because the model has no access to the radargram's internal
  geometry, the output carries no ground truth to validate against.}
  \label{fig:genai-fail}
\end{figure}

\begin{figure*}[t]
\centering
\includegraphics[width=0.9\linewidth,height=5.5cm]{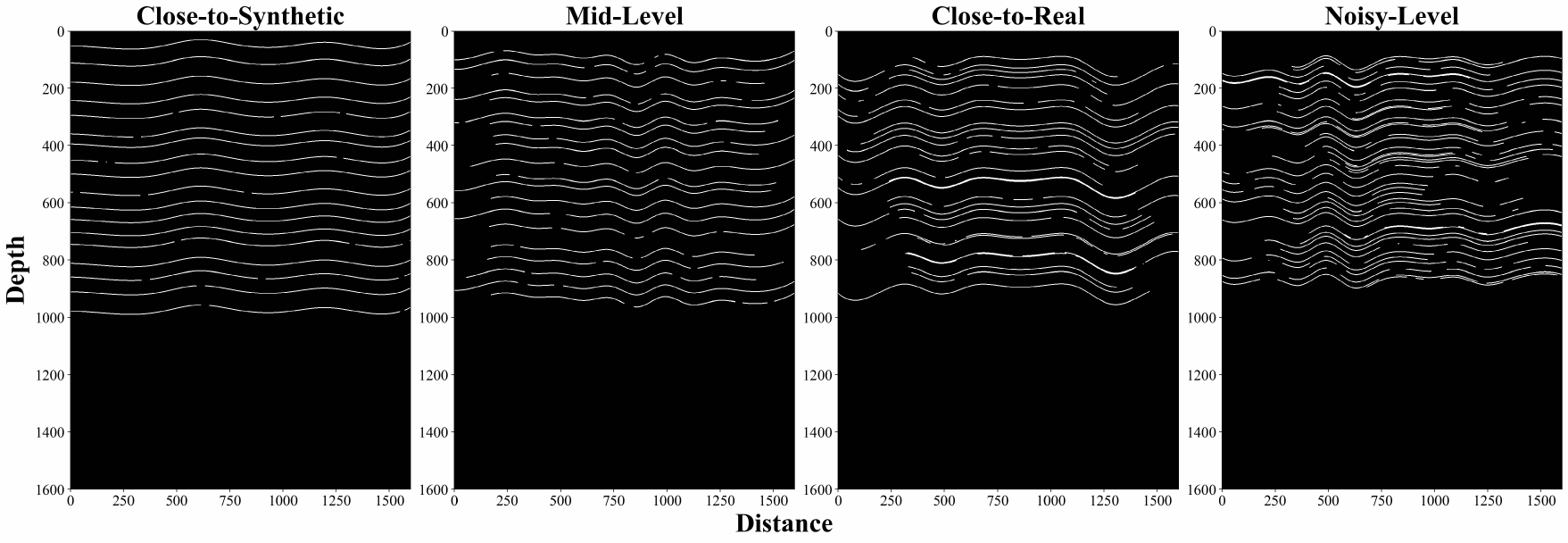}
\caption{SyntheIce Synthetic Radargram Complexity levels of englacial layers.}
\label{fig:samples}
\end{figure*}

\section{Synthetic Radargram Generation}
\label{sec:synthetic}
Scoring a tracing method on real radargrams is bounded by the quality of the expert labels it is compared against: missed layers, gaps, and inconsistent picks make it hard to separate algorithm error from label error. To remove this ambiguity we build \textit{SyntheIce} (Algorithm~\ref{alg:syntheice}), a controlled generator that emits radargrams with exact per-layer ground truth, so
a result can be measured against true geometry rather than a second annotation. Each radargram is generated on a grid of $H \times W = 1600 \times 1600$ pixels, where the vertical axis $H$ is depth (rows) and the horizontal axis $W$ is along-track distance (columns), matching the frame size of a single CReSIS radargram~\cite{paden2014mcords}. It holds up to $K$ layers whose base depths $d_1,\dots,d_K$ are stacked with a random spacing. A shared deformation field $u$, a smooth cubic spline, is added to every layer so they bend together, with a small per-layer offset, while a minimum separation $\delta$ stops consecutive layers from crossing. The visibility mask $\mathit{vis}$ keeps the columns that survive random breaks, turning a continuous layer into fragments, and the tilt angle $\theta$ shifts each column downward to mimic a sloping surface. Wide radargrams use a larger $2000 \times 3000$ grid, cropped along-track into four equal $2000 \times 750$ tiles for checking layer consistency across tiles with XCheck~\cite{xcheck}. The outputs are the binary mask $M$, the per-pixel layer-identity map $L$, the one-pixel skeleton $S$, and the fragment table $F$.

\textit{Why Not Generative AI?}
Before building a controlled generator, we tested whether current image
generation models could produce radargram annotations directly. We gave them
expert radargram and mask pairs from the CReSIS data~\cite{macgregor2015} as
in-context examples and asked, in a one-shot setting, for a binary englacial
layer mask for a new radargram, with the prompt specifying thin, continuous,
smoothly curving layers and gaps where the signal is weak. We tried Google's
Nano Banana models (Gemini~2.5 Flash Image and the Pro tier built on
Gemini~3 Pro Image)~\cite{google2025nanobanana} and OpenAI's
GPT-Image-1~\cite{openai2025gptimage}.

\begin{algorithm}[t]
\SetAlgoLined
\DontPrintSemicolon
\caption{SyntheIce: synthetic radargram generation}
\label{alg:syntheice}

\KwIn{grid $H{\times}W$; layer count $K$; spacing, break, jitter parameters; tilt $\theta$}
\KwOut{mask $M$; layer-id map $L$; per-layer skeleton $S$; fragment table $F$}

$d_1 \leftarrow$ surface depth;\quad
$\mathit{depths} \leftarrow [\,d_1\,]$\;

\For{$i \leftarrow 2$ \KwTo $K$}{
    $d_i \leftarrow d_{i-1} + \mathrm{Uniform(spacing)}$\;

    \lIf{$d_i > H$}{\textbf{break}}

    append $d_i$ to $\mathit{depths}$\;
}

$u \leftarrow$ zero-mean cubic spline through random knots, scaled by jitter\;

$\mathit{prev} \leftarrow \varnothing$\;

\ForEach{$d_i \in \mathit{depths}$}{
    $c \leftarrow d_i + u + \mathrm{offset}_i$\;

    \lIf{$\mathit{prev} \neq \varnothing$}{
        $c \leftarrow \max(c,\ \mathit{prev}+\delta)$
    }

    $c \leftarrow \mathrm{clip}(c, 0, H{-}1)$\;

    $\mathit{vis} \leftarrow$
    \textsc{AssignBreaks}($c$, break params)\;

    optionally trim ends of $\mathit{vis}$ (partial layer)\;

    draw $c$ into $M$ and $L$ with identity $i$ where $\mathit{vis}$\;

    $\mathit{prev} \leftarrow c$\;
}

\lIf{\textnormal{wide}}{
    $(M,L) \leftarrow$ \textsc{Tilt}($M,L,\theta)$
}

$F \leftarrow$ \textsc{ExtractFragments}($L$);\quad
$S \leftarrow$ \textsc{SkeletonizePerLayer}($L$)\;

\Return $M, L, S, F$\;

\end{algorithm}

\begin{table*}[t]
\centering
\caption{Comparison of annotation statistics between real Greenland radar datasets and the four SyntheIce categories.}
\label{tab:dataset-comparison}

\renewcommand{\arraystretch}{1.1}
\setlength{\tabcolsep}{4pt}

\begin{tabular*}{\textwidth}{@{\extracolsep{\fill}}|
>{\centering\arraybackslash}p{2.5cm}|
>{\centering\arraybackslash}p{2.3cm}|
>{\centering\arraybackslash}p{2.2cm}|
>{\centering\arraybackslash}p{2.2cm}|
>{\centering\arraybackslash}p{1.65cm}|
>{\centering\arraybackslash}p{1.65cm}|
>{\centering\arraybackslash}p{1.65cm}|
>{\centering\arraybackslash}p{1.65cm}|}
\hline

&
\multicolumn{3}{c|}{\textbf{Real Greenland Radar Datasets}}
&
\multicolumn{4}{c|}{\textbf{SyntheIce Categories (This Work)}}
\\
\cline{2-8}

\textbf{Statistic}
&
\makecell[c]{\small\textbf{Snow Radar}\\
\small\textbf{(OIB/SRED)~\cite{ibikunle2025}}}
&
\makecell[c]{\small\textbf{Depth Sounder}\\
\small\textbf{(MacGregor)~\cite{macgregor2015}}}
&
\makecell[c]{\small\textbf{ICE2012}\\
\small\textbf{(Rahnemoonfar)}\\
\small\textbf{~\cite{varshney2021}}}
&
\makecell[c]{\rule{0pt}{2.6em}\textbf{Close-to}\\\textbf{Synthetic}}
&
\makecell[c]{\rule{0pt}{2.6em}\textbf{Mid}\\\textbf{Level}}
&
\makecell[c]{\rule{0pt}{2.6em}\textbf{Close-to}\\\textbf{Real}}
&
\makecell[c]{\rule{0pt}{2.6em}\textbf{Noisy}\\\textbf{Level}}
\\
\hline

Layers per radargram
& 15--25
& $\sim$10+
& 12--25
& \textbf{15--19}
& \textbf{18--21}
& \textbf{23--26}
& 35--47
\\
\hline

Fragments per layer
& 1.4--1.8
& 1.0--1.5
& 1.6--2.0
& 1.5--3.0
& 2.0--4.0
& 3.0--5.0
& 3.5--5.0
\\
\hline

Fragment length(px)
& 200--600
& 300--800
& 250--500
& 1001--1769
& 765--1418
& 541--1069
& 442--844
\\
\hline

Annotation width(px)
& 1
& 1
& 1
& \textbf{1}
& \textbf{1}
& \textbf{1}
& \textbf{1}
\\
\hline

Terrain tilt ($^\circ$)
& 0--12
& 0--8
& 0--10
& 1.0--2.8
& 1.1--2.9
& 1.2--3.0
& 1.5--3.0
\\
\hline

Depth variability(px)
& 10--30
& 20--50
& 15--35
& 15--43
& \textbf{14--49}
& 15--46
& \textbf{14--44}
\\
\hline

\end{tabular*}
\end{table*}

The results were not usable for three reasons. First, the outputs were not
faithful: instead of tracing the layers present in the input, the models
hallucinated layers that do not exist and dropped real ones
(Figure~\ref{fig:genai-fail}). Second, layer annotation is a spatial task rather
than an image-generation one, and these models have no access to the internal
geometry of the radargram, so their outputs carry no ground truth to validate
against, which is exactly the supervision a tracing method needs. Third, the
tools are rate-limited in a way that makes dataset-scale use impractical: the
free tiers stop after two or three generations, and the paid tier caps out at
roughly six to eight images per day, after which quality degrades into the
hallucinations already described. A controlled, code-based generator with exact
ground truth therefore replaces these tools, as the rest of this section
describes.
 
\textit{Layer Geometry and Deformation.}
A radargram is generated on a grid of height $H$ (depth) and width $W$ (along-track distance). Layer depths are placed from the surface downward: the first layer starts near a chosen surface depth, and each deeper layer is offset from the one above by a spacing drawn at random from a configured range, until the maximum depth is reached. This keeps layers ordered and physically plausible. Real internal layers fold together because they share a flow history, so SyntheIce applies one deformation field to all layers at once. A smooth undulation, built as a cubic spline through a few random control points, is added to every layer with a small per-layer offset on top.
Because the undulation is shared, layers bend together rather than crossing at random, and a non-crossing constraint enforces a minimum separation between consecutive layers so that ground-truth identities never merge.
 
\textit{Fragmentation and Terrain Tilt.}
Detected layers are broken in practice, so SyntheIce introduces controlled discontinuities. Each layer is cut by a random number of gaps of random length, set by a break probability and a length range, and layers can be trimmed at either end to model partial coverage. These breaks are what turn a continuous layer into the fragments that a linking method must later reassemble. For wide radargrams a terrain tilt is applied as a per-column vertical shift, so layers deepen smoothly across the profile as a sloping surface would in a real survey.
 
\textit{Ground-Truth Representation.}
For each radargram, SyntheIce stores a binary layer mask, a per-pixel layer-identity map, and a one-pixel skeleton of each layer that matches the single-pixel convention of expert picks. It also records a fragment table, in which each fragment is a maximal run of consecutive columns where a layer is present, with its column span, row span, and pixel count, and a layer table inking fragments to their parent and the metadata of the generation parameters fror that specific radargram annotations. This supervision drives both the consistency check with XCheck~\cite{xcheck} and the reconstruction scoring. Algorithm~\ref{alg:syntheice} summarizes the procedure.
 
\textit{Dataset Categories.}
We generate four categories of increasing layer count and fragmentation, from
clean profiles (Close-to-Synthetic, Mid-Level, Close-to-Real) to a dense,
heavily broken Noisy-Level, each with a fixed fraction of wide, tilted
radargrams. Table~\ref{tab:dataset-comparison} reports their measured statistics
against real Greenland snow-radar and depth-sounder
datasets~\cite{ibikunle2025,macgregor2015,varshney2021}. The lighter categories
fall within the layer-count and fragmentation ranges of real data, while
Noisy-Level pushes density and breakage past that range on purpose, giving a
stress case the real datasets do not contain. Pixel-based quantities such as
fragment length are not directly comparable across datasets of different
resolution, so they read as within-dataset trends rather than absolute matches.

\section{Methodology}
\label{sec:method}
 
\begin{figure*}[t]
\centering
\includegraphics[width=0.8\linewidth]{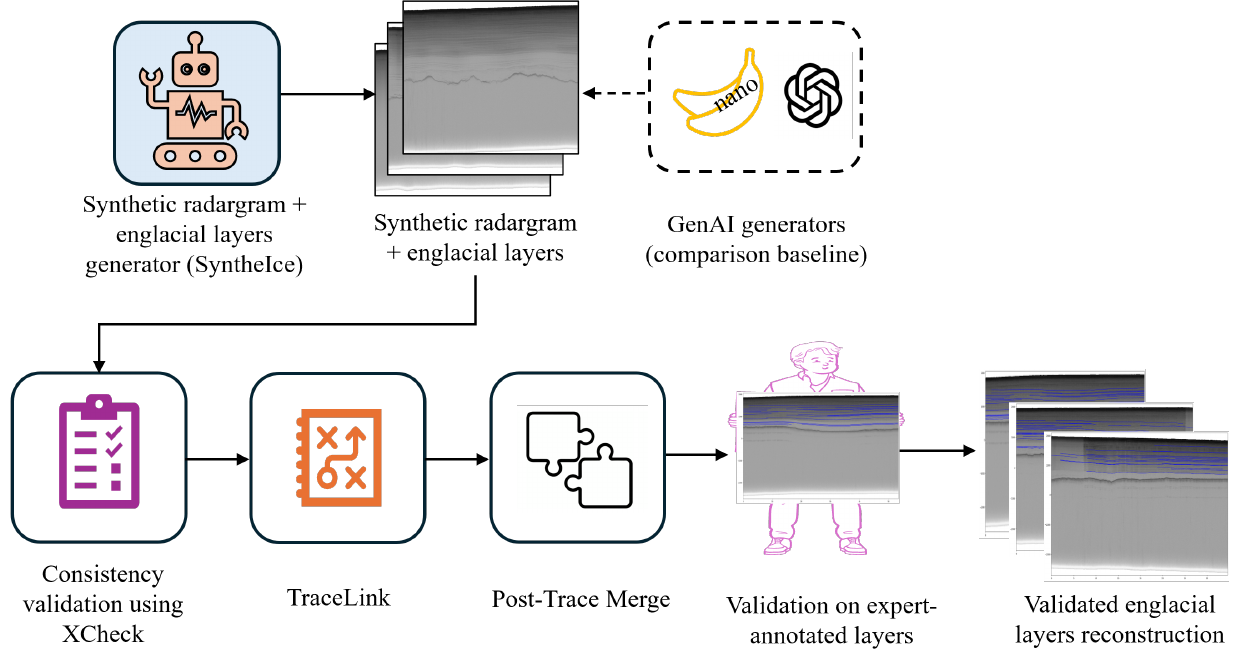}
\caption{End-to-end framework. SyntheIce generates radargrams with exact ground truth; XCheck validates annotation consistency across adjacent tiles; TraceLink reconstructs full-width layers from fragments; the result is validated on real expert-annotated radargrams under controlled fragmentation.}
\label{fig:framework}
\end{figure*}

Figure~\ref{fig:framework} shows the full pipeline of the proposed framework. We generate synthetic radargrams with known ground truth (as shown in Section~\ref{sec:synthetic}), check annotation consistency across neighbouring tiles with XCheck~\cite{xcheck}, reconstruct full-width layers from fragments with \textit{TraceLink} (see Algorithm~\ref{alg:tracelink}), and validate TraceLink on real expert-annotated radargrams under controlled fragmentation.
 
\subsection{Problem Formulation}
A skeletonized radargram yields a set of fragments, where each fragment is a short connected piece of a layer, described geometrically by a column span and a row position. After detection, the parent-layer identity is lost, and the fragments are all that remain. The task is to partition the fragments into layers, assigning every fragment to exactly one layer so that the recovered layers match the true horizons. Formally, let $\mathcal{B}=\{b_1,\dots,b_n\}$ be the set of fragments. We seek a labelling $\phi:\mathcal{B}\rightarrow\{1,\dots,K\}$ that assigns each fragment to exactly one layer, so that fragments of the same true horizon share a label and $K$ matches the number of true layers. TraceLink (Algorithm~\ref{alg:tracelink}) computes $\phi$ directly; it does not fill the gap pixels between fragments. The method maps fragments to layers only. It does not paint pixels into the empty gaps between fragments, which keeps the output faithful to what was actually detected.

\subsection{Consistency Validation using XCheck}
For the wide radargrams, each radargram is cropped into four neighboring tiles. Before reconstruction, XCheck~\cite{xcheck} confirms that the annotations agree between adjacent tiles of the same radargram. XCheck scores layer agreement across tile boundaries to flag inconsistent or spurious picks. The check is applied to
all wide categories of synthetic annotations so that label quality is verified before data enter the pipeline.
 
\subsection{TraceLink: Slope-Aware Fragment Linking}
TraceLink scans fragments from left to right and grows layers greedily. Candidate layers for a fragment are those whose right edge lies within a column window to its left. For each candidate, TraceLink predicts where the layer would be at the fragment's starting column from the layer's current slope, and compares that prediction to the fragment's row. A match is accepted only if the row difference is within a tolerance that widens with the predicted drift, so steeper layers are allowed a larger vertical step. When several candidates qualify, the one closest in row wins, not the one nearest in column. This row-first priority is what stops a fragment from being captured by a horizontally closer but vertically wrong layer. Concretely, for a fragment $b$ and a candidate layer $\ell$
(Algorithm~\ref{alg:tracelink}), the predicted row and its mismatch are
\begin{equation}
  \hat{r}_{\ell} = \ell.\mathrm{row} + \ell.\mathrm{slope}\cdot\Delta\mathrm{col},
  \qquad
  \rho_{\ell} = \lvert\, \hat{r}_{\ell} - b.\mathrm{row} \,\rvert ,
\end{equation}
and $\ell$ accepts $b$ only within an adaptive tolerance that widens with the
predicted drift,
\begin{equation}
  \rho_{\ell} \le \tau + \beta\,\lvert \ell.\mathrm{slope}\rvert\,\Delta\mathrm{col}.
  \label{eq:tol}
\end{equation}
Among the candidates $\mathcal{C}$ that pass, $b$ joins the one closest in row,
with ties broken by the rightmost edge,
\begin{equation}
  \ell^{\ast} = \operatorname*{arg\,max}_{\ell \in \mathcal{C}}\;
  \bigl(-\rho_{\ell},\; \ell.\mathrm{right}\bigr).
\end{equation}
Here $b.\mathrm{row}$ is the fragment's row and $\ell.\mathrm{row}$,
$\ell.\mathrm{slope}$ the layer's current row and slope; $\Delta\mathrm{col}$ is
the column gap between the layer's last fragment and $b$; $\tau$ is the base row
tolerance and $\beta$ the slope factor.
 
Each layer tracks its slope as an exponentially weighted moving average of the row change per column across its fragments. After each assignment the slope is updated as
\begin{equation}
  \ell.\mathrm{slope} \leftarrow (1-\alpha)\,\ell.\mathrm{slope}
  + \alpha\,s_{b},
  \label{eq:ewma}
\end{equation}
where $s_{b}$ is the row-per-column slope from the layer's previous fragment to
$b$ and $\alpha$ is the EWMA weight. So the estimate adapts along the layer while staying stable against a single noisy fragment. The weighted moving average (EWMA) weight '$\alpha$' trades responsiveness against stability. A high $\alpha \;(\rightarrow 1.0)$ follows curvature fast but is easily thrown by a noisy fragment, and a low $\alpha \;(\rightarrow 0.1)$ resists noise but lags on real bends; we set (EWMA) weight $\alpha = 0.5$ to balance the two. The column window and row tolerance are not fixed: they are derived from the spacing and gap statistics of each radargram, so the same method runs on sparse synthetic profiles and dense real ones without manual tuning. Each recovered layer also receives a stable content-based identifier, so the same layer keeps its identity across radargrams.
 
\subsection{Post-Trace Merge}

Greedy linking can still split one layer in two when a gap is wide. A strict merge pass repairs this. It considers pairs of column-disjoint layers and fits a line to each layer's fragments. Two layers are merged only if the gap between them is small relative to their own internal gaps, the first layer's fitted line predicts the start of the second, and their slopes agree. The fit uses a regression over all fragments rather than the reactive moving-average slope, which keeps the decision stable. Merging is bookkeeping only: fragment labels are reassigned and no gap pixels are filled. The pass repeats until no further pair qualifies, after which each layer receives a slope-consistency confidence and the layers are
renumbered top to bottom (the final \textsc{MergeStrict},
\textsc{SlopeConfidence}, and renumbering steps of
Algorithm~\ref{alg:tracelink}).

\subsection{Real-Data Validation}
To test TraceLink on real data without depending on noisy labels, we turn clean expert layers into a controlled problem. From an expert-annotated Greenland radargram by Macgregor et al., 2015 ~\cite{macgregor2015} and the raw radargrams from CReSIS portal ~\cite{cresis2024rds}, we keep only layers that span at least 70\% of the width, since these are continuous enough to act as ground truth. We inject random gaps between 20--40 pixels/width(m) into them to mimic detector fragmentation, run TraceLink on the broken result, and compare the reconstruction against the original layer. Scoring uses intersection-over-union (IoU) between predicted and true layers on equal-width bands, so thin layers are compared fairly. Because the injected gaps are excluded from the reference, the method is not penalized for leaving them empty: the score reflects how correctly fragments are grouped, not whether pixels were invented. We report mean IoU, precision, and F1, with a sensitivity check on the linking tolerance.
 
\begin{algorithm}[t]
\SetAlgoLined
\DontPrintSemicolon
\caption{TraceLink: slope-aware fragment linking}
\label{alg:tracelink}

\KwIn{binary mask $M$; column window $g$; row tolerance $\tau$}
\KwOut{layer map; fragment-to-layer assignment}

$\mathcal{B} \leftarrow$ 
\textsc{ConnectedComponents}(\textsc{Skeletonize}($M$)), 
filtered by width and area\;

sort $\mathcal{B}$ by left edge, then row\;

\ForEach{fragment $b \in \mathcal{B}$}{
    $\mathcal{C} \leftarrow$ layers with right edge in 
    $[\,b.\mathrm{left}-g,\ b.\mathrm{left}\,]$\;

    $\mathit{best} \leftarrow \varnothing$;\quad
    $\mathit{key}^\ast \leftarrow (-\infty, -)$\;

    \ForEach{layer $\ell \in \mathcal{C}$}{
        $\hat{r} \leftarrow 
        \ell.\mathrm{row} + 
        \ell.\mathrm{slope}\cdot\Delta\mathrm{col}$\;

        $\rho \leftarrow |\hat{r} - b.\mathrm{row}|$;\quad
        $\mathit{tol} \leftarrow 
        \tau + \beta\,|\ell.\mathrm{slope}|\,\Delta\mathrm{col}$\;

        \If{$\rho \leq \mathit{tol}$ 
        \textnormal{\textbf{and}} 
        $(-\rho,\ \ell.\mathrm{right}) > \mathit{key}^\ast$}{
            $\mathit{best} \leftarrow \ell$;\quad
            $\mathit{key}^\ast \leftarrow 
            (-\rho,\ \ell.\mathrm{right})$\;
        }
    }

    \eIf{$\mathit{best} \neq \varnothing$}{
        assign $b$ to $\mathit{best}$; 
        update slope by EWMA; 
        update right edge and row\;
    }{
        start a new layer from $b$\;
    }
}

relabel fragments to layers\;

\textsc{MergeStrict} (line-fit gated);\quad
\textsc{SlopeConfidence};\quad
renumber top to bottom\;

\Return layer map, assignment\;

\end{algorithm}

\section{Experiments and Results}
\label{sec:experiments}
 
We evaluate the framework in three steps that follow Figure~\ref{fig:framework} such as annotation consistency on synthetic data with XCheck, fragment linking on synthetic data with TraceLink, and fragment linking on real expert annotations \cite{macgregor2015}. The first step confirms that the ground
truth is trustworthy, the second measures reconstruction against exact labels,
and the third tests transfer to real radargrams.
 
\subsection{Experimental Setup}
\label{sec:exp-setup}
 
\textit{Data.} All synthetic radargrams were obtained from SyntheIce across the four categories shown in Table~\ref{tab:dataset-comparison}. Experiment~1 uses 150 wide radargrams, Experiment~2 uses 400 radargrams with 100 per category, and Experiment~3 uses expert-annotated Greenland radargrams from airborne surveys~\cite{macgregor2015}.

\begin{figure*}[tbp]
  \centering
  \includegraphics[width=\textwidth]{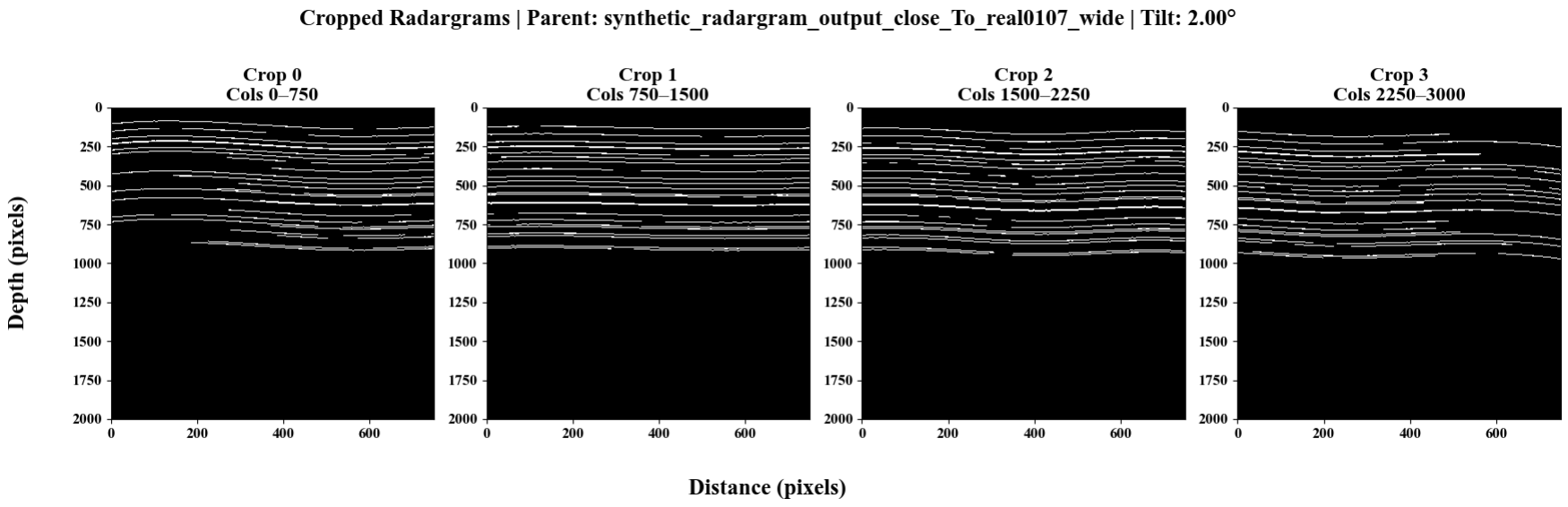}
  \caption{A wide radargram (Close-to-Real category, terrain tilt $2.6^\circ$)
  cropped into four adjacent tiles. XCheck validates layer agreement across
  the shared tile boundaries against the ground truth of the original
  uncropped radargram.}
  \label{fig:crop}
\end{figure*}

\textit{Metrics.} We report precision, recall, F1, and
intersection-over-union (IoU). A predicted
layer counts as correct when its IoU with a ground-truth layer passes a fixed threshold; mean IoU is then averaged over the matched layers. For the real-data evaluation, we report the same metrics.
 
\textit{Implementation.} The column window and row tolerance are derived per
radargram from its own spacing and gap statistics, so no parameter is
hand-tuned (Section~\ref{sec:method}). Gap injection uses a fixed random seed
for reproducibility.
 
\subsection{Layers Annotation Consistency with XCheck}
\label{sec:exp1}
 
This experiment checks that SyntheIce annotations are internally consistent and that XCheck~\cite{xcheck} catches disagreement where layers meet. Each cropped wide radargrams stored 600 tiles as mentioned in (Section~\ref{sec:synthetic}),with all meta data of their englacial layers internal structure. Figure~\ref{fig:crop} shows one such split.
XCheck then compares layers across the shared boundaries of neighboring tiles, and its output is checked against the ground truth of the original uncropped radargram, so a correct result means the per-tile annotations agree exactly where the tiles join.
 
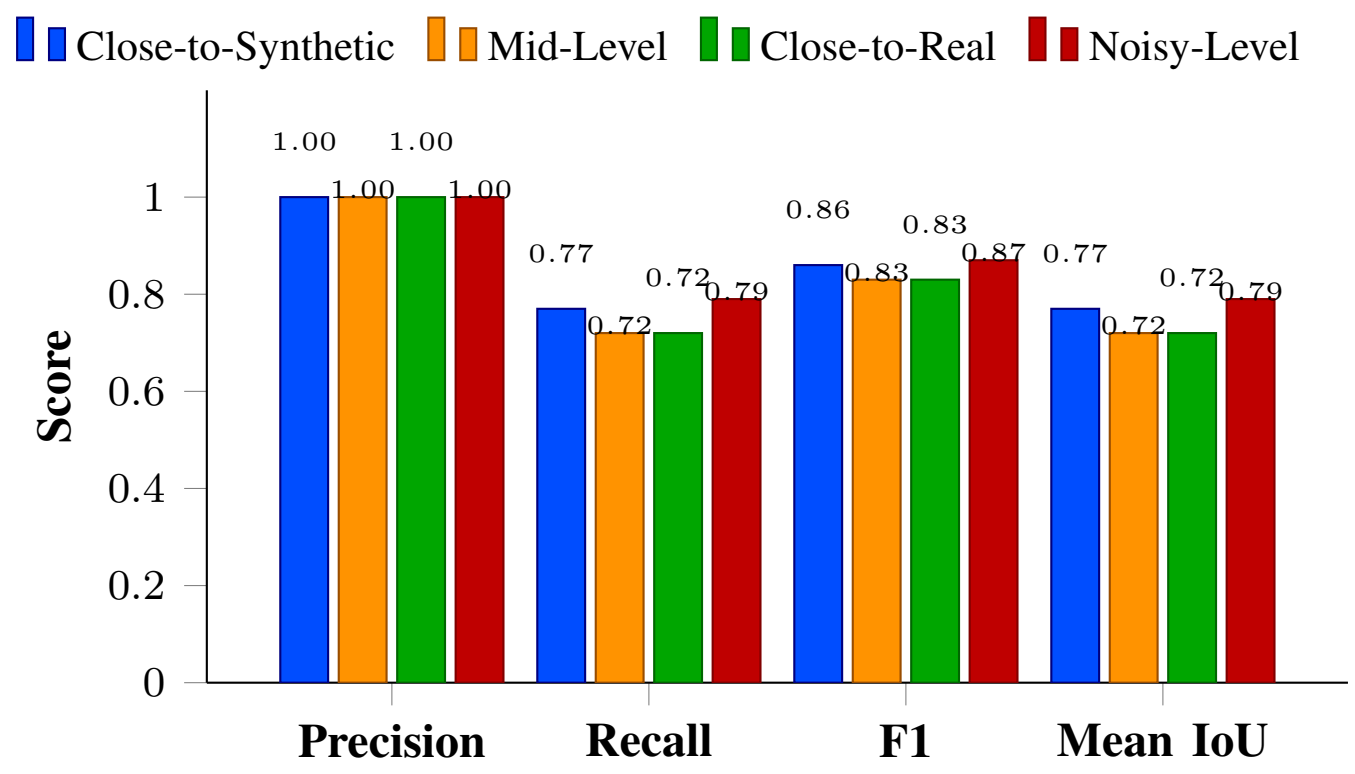
\begin{figure}[t]
  \centering
  \begin{tikzpicture}
    \begin{axis}[
      ybar,
      width=0.90\linewidth,
      height=5.5cm,
      enlarge x limits=0.24,
      bar width=9pt,
      ymin=0, ymax=1.22,
      ytick={0,0.2,0.4,0.6,0.8,1.0},
      yticklabel={\pgfmathprintnumber[fixed,precision=1]{\tick}},
      ylabel={\textbf{Score}},
      ylabel style={font=\small},
      symbolic x coords={Precision,Recall,F1,Mean IoU},
      x=1.7cm,
      xtick=data,
      xticklabel style={font=\small\bfseries},
      yticklabel style={font=\footnotesize\bfseries},
      nodes near coords,
      nodes near coords style={
        font=\fontsize{5}{6}\selectfont\bfseries,
        text=black,
        anchor=south,
        inner sep=0.6pt,
        /pgf/number format/.cd,
        fixed,
        fixed zerofill,
        precision=2,
        /tikz/.cd,
      },
      legend style={
        at={(0.4,1.00)},
        anchor=south,
        legend columns=4,
        font=\footnotesize,
        draw=none,
        /tikz/every even column/.append style={column sep=4pt}
      },
      tick align=outside,
      tick pos=left,
      axis x line*=bottom,
      axis y line*=left,
    ]
 
    % Close-to-Synthetic
    \addplot+[fill=blue!70!cyan, draw=blue!50!black,
              every node near coord/.append style={yshift=8pt}] coordinates {
      (Precision, 1.00) (Recall, 0.77) (F1, 0.86) (Mean IoU, 0.77)
    };

    % Mid-Level
    \addplot+[fill=orange!85!yellow, draw=orange!60!black,
              every node near coord/.append style={yshift=-1pt}] coordinates {
      (Precision, 1.00) (Recall, 0.72) (F1, 0.83) (Mean IoU, 0.72)
    };

    % Close-to-Real
    \addplot+[fill=green!65!black, draw=green!40!black,
              every node near coord/.append style={yshift=8pt}] coordinates {
      (Precision, 1.00) (Recall, 0.72) (F1, 0.83) (Mean IoU, 0.72)
    };

    % Noisy-Level
    \addplot+[fill=red!75!black, draw=red!50!black,
              every node near coord/.append style={yshift=-1pt}] coordinates {
      (Precision, 1.00) (Recall, 0.79) (F1, 0.87) (Mean IoU, 0.79)
    };
 
    \legend{Close-to-Synthetic, Mid-Level, Close-to-Real, Noisy-Level}
    \end{axis}
  \end{tikzpicture}

  \caption{XCheck layers consistency scores on the 150 synthetic wide radargrams cropped into 600 by complexity levels of SyntheIce referred as categories. Precision, recall, F1 and Mean IoU for all categories.}
  \label{fig:xcheck}
\end{figure}

Figure~\ref{fig:xcheck} reports the scores by category. Precision is at or near 1.0 everywhere, from 0.999 to 1.000, so XCheck almost never marks a correct layer as inconsistent. Recall ranges from 0.72 to 0.79 and F1 from 0.83 to 0.87, with the denser Mid-Level and Close-to-Real categories slightly lower because more layers meet at each boundary. IoU equals recall in every
category with almost no false positives.

\subsection{Fragment Linking on Synthetic Data}
\label{sec:exp2}
 
Here we measure TraceLink against exact ground truth on 400 radargrams (100 per category), scoring the traced layers against the SyntheIce layer-identity ground-truth map as explained in the (Section~\ref{sec:method}). 

\begin{figure}[tbp]
  \centering
  \includegraphics[width=\linewidth,height=5cm]{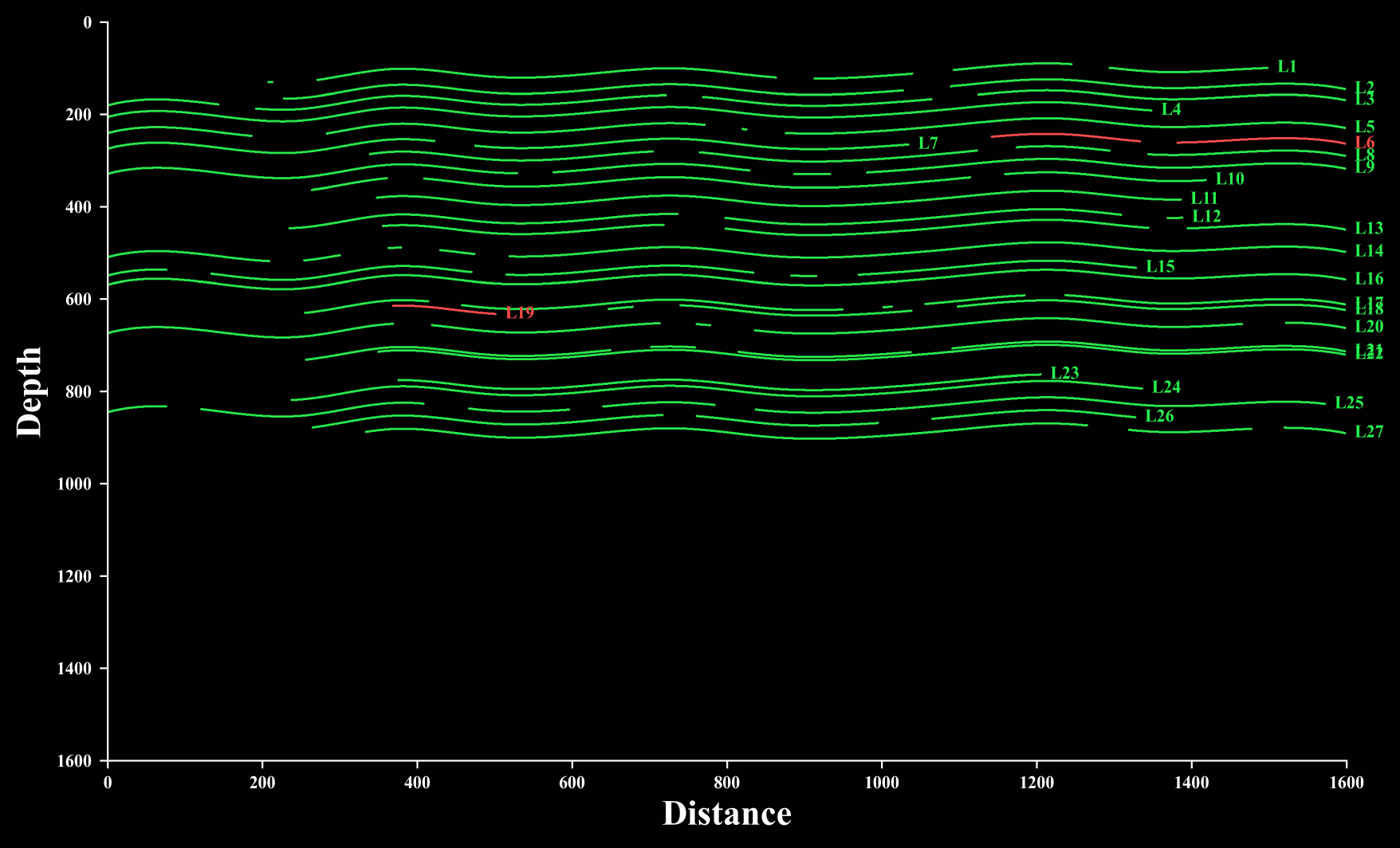}
  \caption{TraceLink applied on (synthetic\_radargram\_close\_To\_real067.h5) id. Fragments that are correctly linked and mapped to their corresponding ground-truth layers are shown in \textcolor{green!50!black}{Green}, whereas fragments associated with incorrect layer assignments are shown in \textcolor{red}{red}. TraceLink results Precision=0.93, Recall=1.00, F1=0.96 and Mean IoU=0.94.}
  \label{fig:tl-syn-example}
\end{figure}

\begin{figure}[tbp]
  \centering
  \begin{tikzpicture}
    \begin{axis}[
      ybar,
      width=0.90\linewidth,
      height=5.5cm,
      enlarge x limits=0.24,
      bar width=9pt,
      ymin=0, ymax=1.22,
      ytick={0,0.2,0.4,0.6,0.8,1.0},
      yticklabel={\pgfmathprintnumber[fixed,precision=1]{\tick}},
      ylabel={\textbf{Score}},
      ylabel style={font=\small},
      symbolic x coords={Precision,Recall,F1,Mean IoU},
      x=1.7cm,
      xtick=data,
      xticklabel style={font=\small\bfseries},
      yticklabel style={font=\footnotesize\bfseries},
      nodes near coords,
      nodes near coords style={
        font=\fontsize{5}{6}\selectfont\bfseries,
        text=black,
        anchor=south,
        inner sep=0.6pt,
        /pgf/number format/.cd, fixed, fixed zerofill, precision=2, /tikz/.cd,
      },
      legend style={
        at={(0.4,1.00)}, anchor=south,
        legend columns=4,
        font=\footnotesize,
        draw=none,
        /tikz/every even column/.append style={column sep=4pt}
      },
      tick align=outside,
      tick pos=left,
      axis x line*=bottom,
      axis y line*=left,
    ]
    % Values at IoU matching threshold theta_IoU = 0.50
    % Staggered labels: alternating up/down so adjacent values don't collide
    \addplot+[fill=blue!70!cyan, draw=blue!50!black,
              every node near coord/.append style={yshift=8pt}] coordinates {
      (Precision, 0.96) (Recall, 1.00) (F1, 0.98) (Mean IoU, 0.99)
    };
    \addplot+[fill=orange!85!yellow, draw=orange!60!black,
              every node near coord/.append style={yshift=-1pt}] coordinates {
      (Precision, 0.95) (Recall, 1.00) (F1, 0.97) (Mean IoU, 0.99)
    };
    \addplot+[fill=green!65!black, draw=green!40!black,
              every node near coord/.append style={yshift=8pt}] coordinates {
      (Precision, 0.84) (Recall, 0.95) (F1, 0.89) (Mean IoU, 0.93)
    };
    \addplot+[fill=red!75!black, draw=red!50!black,
              every node near coord/.append style={yshift=-1pt}] coordinates {
      (Precision, 0.82) (Recall, 0.89) (F1, 0.85) (Mean IoU, 0.90)
    };

    \legend{Close-to-Synthetic, Mid-Level, Close-to-Real, Noisy-Level}
    \end{axis}
  \end{tikzpicture}
  \caption{TraceLink reconstruction scores on 400 synthetic radargrams
  (100 per category) at IoU matching threshold $\theta_{\mathrm{IoU}}=0.50$:
  precision, recall, F1, and mean IoU (macro-averaged).}
  \label{fig:tl-syn-bars}
\end{figure}
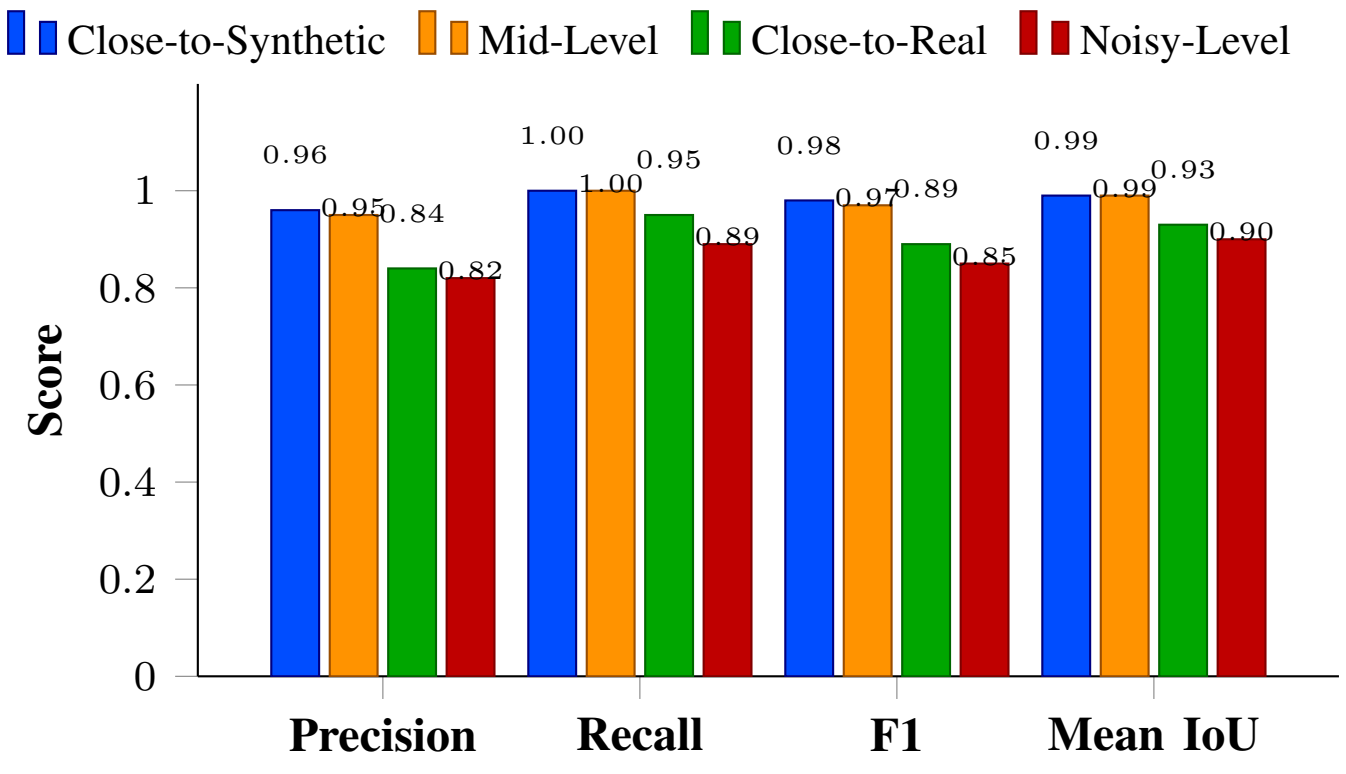

Figure~\ref{fig:tl-syn-example} shows one result, where broken fragments are grouped into continuous layers without filling the gaps between them. Figure~\ref{fig:tl-syn-bars} reports the per-category scores at
$\theta_{\mathrm{IoU}} = 0.50$. Recall stays at or above 0.89 while precision and mean IoU decline on denser data, from 0.96 to 0.82 in precision and F1 between 0.85 and 0.98. The strict merge pass reduces the predicted-layer count by 1--2\% per category (e.g., $5147 \rightarrow 5085$ on Noisy-Level, $2075 \rightarrow 2032$ on Close-to-Synthetic), trimming over splitting from greedy linking. The result of threshold ranges in (Table~\ref{tab:iou-sensitivity}) confirms this: F1 is score is nearly unchanged from 0.20 to 0.70 on the easier categories, and although it drops on Close-to-Real and Noisy-Level, the mean IoU of the matched layers rises at every step, reaching 0.96 at IoU~$\geq$~0.70. The retained layers are therefore high-overlap traces, and TraceLink (Algorithm~\ref{alg:tracelink}) reconstructs layers reliably across all four complexity levels.
 
%\begin{figure}[tbp]
%  \centering
%  \includegraphics[width=\linewidth]{Figures/TRACElink_synthetic.pdf}
%  \caption{TraceLink reconstruction scores on 400 synthetic radargrams
%  (100 per category): precision, recall, F1, and mean IoU. Recall stays near
%  1.0 across categories; precision and IoU decline on the denser Close-to-Real
%  and Noisy-Level data.}
%  \label{fig:tl-syn-bars}
%\end{figure}

\subsection{Fragment Linking on Real Expert Annotations}
\label{sec:exp3}
 
\begin{figure}[tbp]
  \centering
  \includegraphics[width=\linewidth,height=5cm]{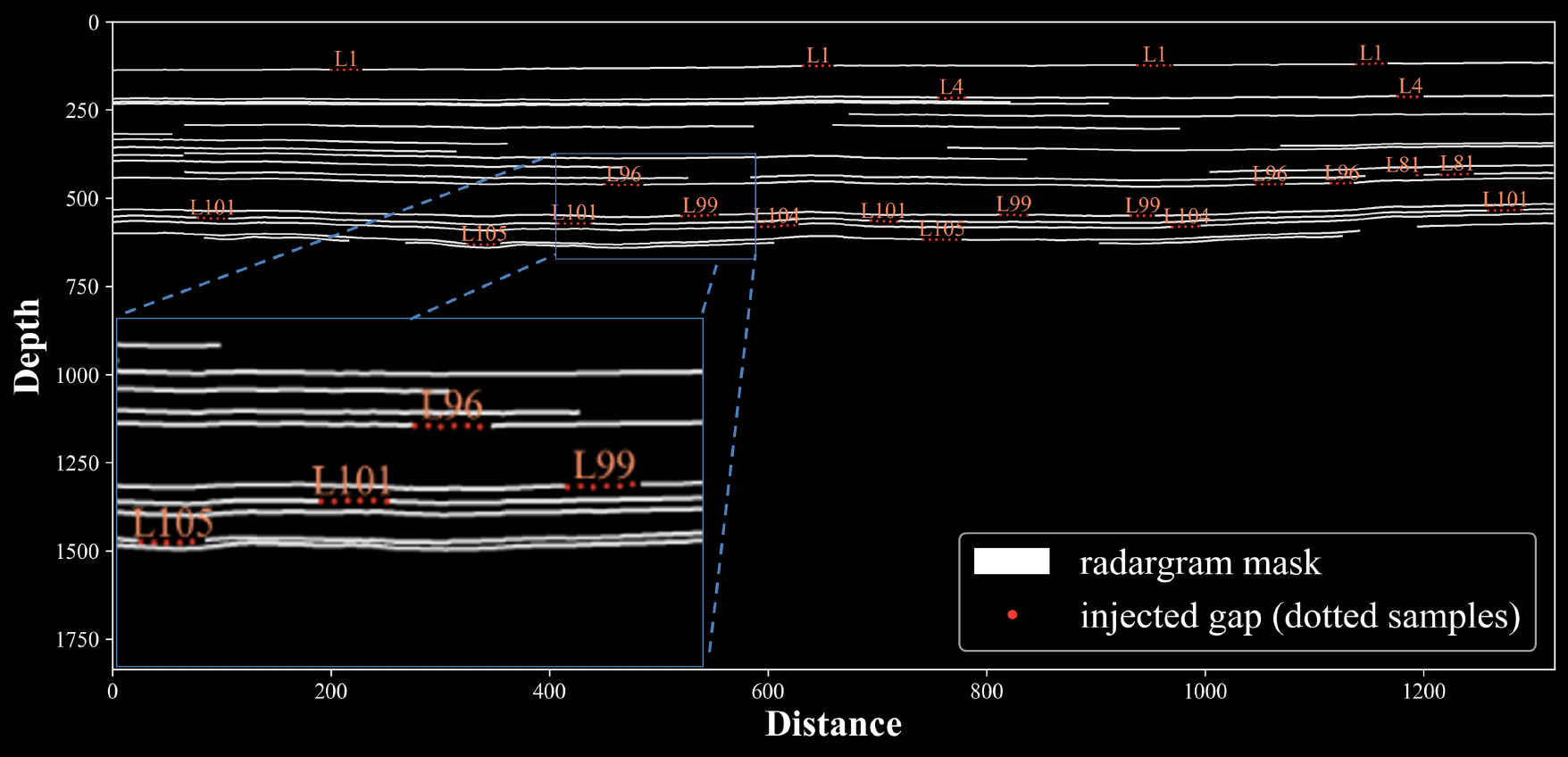}
  \caption{Controlled fragmentation on a real (Data\_20120516\_01\_084) expert-annotated radargram.Layers covering at least 70\% of the width are selected, and random gaps
  (dotted samples) are injected to imitate detector breakage. The inset shows
  the gaps at full resolution. Removed gaps are excluded from the reference
  during scoring.}
  \label{fig:gaps}
\end{figure}

\begin{figure*}[tbp]
  \centering
  \includegraphics[width=\linewidth]{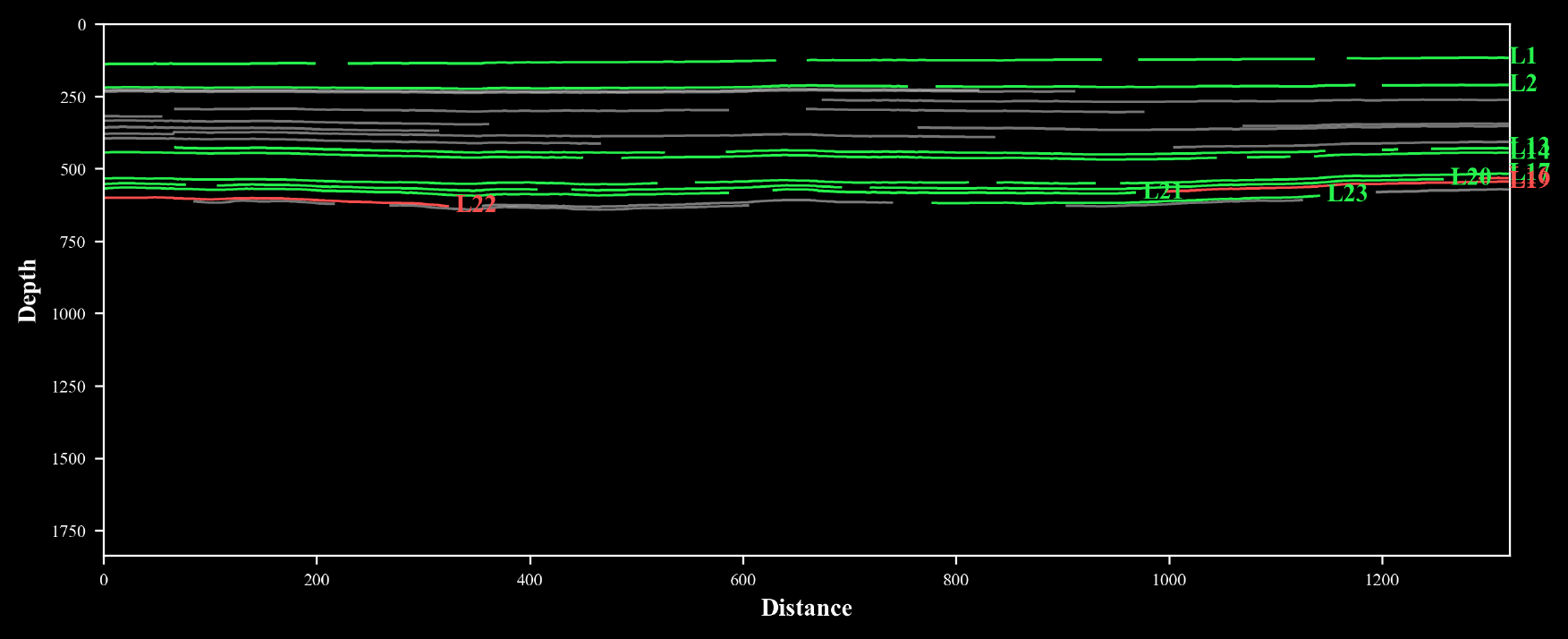}
  \caption{TraceLink on a real radargram (Data\_20120516\_01\_084) after gap
  injection: The fragments that are \textcolor{green!50!black}{Green} are correctly traced and reconstructed to the original full layers versus ground truth, whereas fragments associated with incorrect layer assignments are shown in \textcolor{red}{Red}. The layers that are in (Grey) are the layers we are not using. TraceLink had 100\% layer recovery on
  this profile.}
  \label{fig:tl-real-example}
\end{figure*}

The final experiment tests whether TraceLink transfers to real data. Real expert
labels are clean but carry no controlled fragmentation, so we build a controlled
problem from them: from each expert-annotated radargram we keep the layers that
span at least 70\% of the width, since these are continuous enough to serve as
ground truth, remove speckle with a connected-component clean-up, and inject
random gaps to imitate detector breakage (Figure~\ref{fig:gaps}). We run
TraceLink on the broken layers and compare the reconstruction against the
original layer (Figure~\ref{fig:tl-real-example}). Scoring uses IoU on
equal-width bands so thin layers are judged fairly, and the injected gaps are
excluded from the reference, so the method is measured on how well it groups
fragments, not on whether it fills gaps it was never meant to fill. This protocol
is rigorous because the answer is known (gaps are removed from layers that were
continuous to begin with), the metric compares against the original expert layer
rather than a second annotation, and the test is reproducible and free of
hand-tuning, since the gap pattern is seeded and the linking parameters are
derived from each radargram.

\begin{figure}[tbp]
  \centering
  \begin{tikzpicture}
    \begin{axis}[
      ybar,
      width=0.9\linewidth,
      height=5.5cm,
      enlarge x limits=0.16,
      bar width=28pt,
      ymin=0, ymax=1.12,
      ytick={0,0.2,0.4,0.6,0.8,1.0},
      yticklabel={\pgfmathprintnumber[fixed,precision=1]{\tick}},
      ylabel={\textbf{Score}},
      ylabel style={font=\small},
      symbolic x coords={Precision,Recall,F1,Mean IoU},
      xtick={Precision,Recall,F1,Mean IoU},
      xticklabel style={font=\small\bfseries},
      yticklabel style={font=\footnotesize\bfseries},
      tick align=outside,
      tick pos=left,
      axis x line*=bottom,
      axis y line*=left,
    ]

    % Precision — diagonal lines
    \addplot+[ybar, bar width=28pt, bar shift=0pt,
        draw=black, line width=0.6pt,
        fill=gray!25,
        postaction={
          pattern={Lines[angle=45, distance=3pt, line width=1.4pt]},
          pattern color=black!25
        }]
    coordinates {(Precision, 0.78)};

    % Recall — crosshatch diagonal
    \addplot+[ybar, bar width=28pt, bar shift=0pt,
        draw=black, line width=0.6pt,
        fill=gray!25,
        postaction={
          pattern={Hatch[angle=45, distance=3pt, line width=1.4pt]},
          pattern color=black!25
        }]
    coordinates {(Recall, 0.93)};

    % F1 — reverse diagonal lines
    \addplot+[ybar, bar width=28pt, bar shift=0pt,
        draw=black, line width=0.6pt,
        fill=gray!25,
        postaction={
          pattern={Lines[angle=135, distance=3pt, line width=1.4pt]},
          pattern color=black!25
        }]
    coordinates {(F1, 0.84)};

    % Mean IoU — dots
    \addplot+[ybar, bar width=28pt, bar shift=0pt,
        draw=black, line width=0.3pt,
        fill=gray!25,
        postaction={
          pattern={Dots[distance=4pt, radius=1.4pt]},
          pattern color=black!25
        }]
    coordinates {(Mean IoU, 0.82)};

    \node[font=\small\bfseries] at (axis cs:Precision,0.78) [above=2pt] {0.78};
    \node[font=\small\bfseries] at (axis cs:Recall,0.93)    [above=2pt] {0.93};
    \node[font=\small\bfseries] at (axis cs:F1,0.84)        [above=2pt] {0.84};
    \node[font=\small\bfseries] at (axis cs:{Mean IoU},0.82)[above=2pt] {0.82};

    \end{axis}
  \end{tikzpicture}
  \caption{TraceLink reconstruction validation protocol scores across the 400
    real expert-annotated CReSiS Greenland radargrams at IoU matching threshold $\theta_{\mathrm{IoU}}=0.50$: precision, recall, F1,
    and mean IoU.}
  \label{fig:tl-real-bars}
\end{figure}
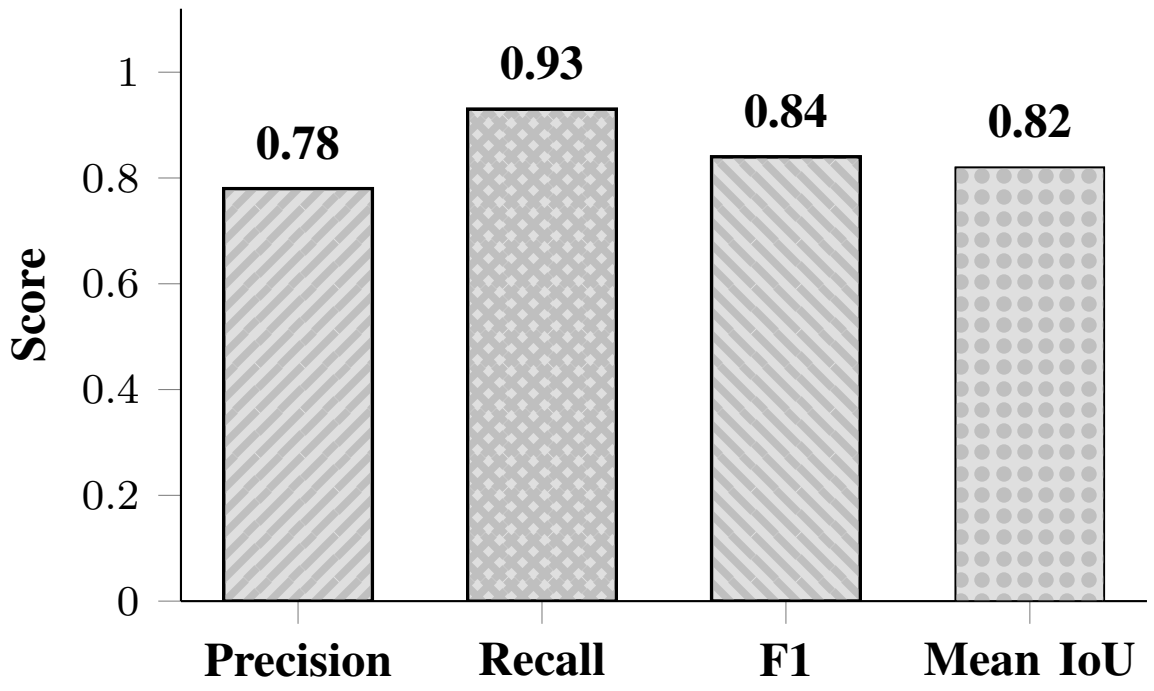

Figure~\ref{fig:tl-real-bars} reports the scores at $\theta_{\mathrm{IoU}} =
0.50$. Recall is high at 0.93, so TraceLink recovers almost every real layer,
while precision (0.78), F1 (0.84), and mean IoU (0.82) are lower than on the
hardest synthetic category, as expected since real layers are thinner, wavier,
and more irregularly broken than the generated ones; the single radargram in
Figure~\ref{fig:tl-real-example} reaches 100\% layer recovery. The strict merge
pass has the same small corrective effect as on synthetic data, reducing the
predicted-layer count from $14024 \rightarrow 13677$ (347 merges, a $2.47\%$
reduction). A threshold sweep (Table~\ref{tab:iou-sensitivity-real}) confirms the
result is robust: from 0.20 to 0.70, precision, recall, and F1 fall only
gradually while the mean IoU of the matched layers rises from 0.808 to 0.838,
showing that TraceLink's real-data reconstructions are high-overlap traces rather
than loose matches that survive only a lenient threshold.

\begin{table*}[t]
\centering
\caption{Sensitivity of TraceLink to the different IoU matching threshold on the 400 synthetic set, about 100 radargrams per category. Scores are stable on the easier categories; on the denser ones F1 drops with the threshold while the mean IoU of matched layers rises, showing the retained layers are high-fidelity traces.}
\label{tab:iou-sensitivity}

\renewcommand{\arraystretch}{1.2}
\setlength{\tabcolsep}{6pt}

\begin{tabular*}{\textwidth}{@{\extracolsep{\fill}}|l|cccc|cccc|cccc|cccc|}
\hline

&
\multicolumn{4}{c|}{\textbf{IoU $\geq$ 0.20}}
&
\multicolumn{4}{c|}{\textbf{IoU $\geq$ 0.30}}
&
\multicolumn{4}{c|}{\textbf{IoU $\geq$ 0.50}}
&
\multicolumn{4}{c|}{\textbf{IoU $\geq$ 0.70}}
\\
\cline{2-17}

\shortstack{\textbf{SyntheIce}\\\textbf{Category}}
& \textbf{P} & \textbf{R} & \textbf{F1} & \textbf{mIoU}
& \textbf{P} & \textbf{R} & \textbf{F1} & \textbf{mIoU}
& \textbf{P} & \textbf{R} & \textbf{F1} & \textbf{mIoU}
& \textbf{P} & \textbf{R} & \textbf{F1} & \textbf{mIoU}
\\
\hline\hline

Close-to-Synthetic
& 0.96 & 1.00 & 0.98 & 0.99
& 0.96 & 1.00 & 0.98 & 0.99
& 0.96 & 1.00 & 0.98 & 0.99
& 0.95 & 0.98 & 0.96 & 1.00
\\
\hline

Mid-Level
& 0.95 & 1.00 & 0.97 & 0.98
& 0.95 & 1.00 & 0.97 & 0.98
& 0.95 & 1.00 & 0.97 & 0.99
& 0.93 & 0.97 & 0.95 & 0.99
\\
\hline

Close-to-Real
& 0.88 & 1.00 & 0.93 & 0.91
& 0.88 & 1.00 & 0.93 & 0.91
& 0.84 & 0.95 & 0.89 & 0.93
& 0.77 & 0.87 & 0.81 & 0.96
\\
\hline

Noisy-Level
& 0.91 & 0.99 & 0.95 & 0.85
& 0.90 & 0.98 & 0.94 & 0.86
& 0.82 & 0.89 & 0.85 & 0.90
& 0.70 & 0.76 & 0.73 & 0.96
\\
\hline

\end{tabular*}
\end{table*}

\begin{table}[t]
\centering
\caption{Sensitivity of TraceLink to the IoU different matching threshold
$\theta_{\mathrm{IoU}}$ on the real expert-annotated CReSIS Greenland
radargrams. As the threshold tightens, the mean IoU of the matched layers rises, indicating that the retained layers are high-overlap reconstructions.}
\label{tab:iou-sensitivity-real}

\renewcommand{\arraystretch}{1.2}
\setlength{\tabcolsep}{6pt}

\begin{tabular*}{\columnwidth}{@{\extracolsep{\fill}}|c|c|c|c|c|}
\hline

\textbf{$\boldsymbol{\theta}_{\boldsymbol{\mathrm{IoU}}}$}
& \textbf{Precision}
& \textbf{Recall}
& \textbf{F1}
& \textbf{Mean IoU}
\\
\hline\hline

\textbf{0.20} & 0.828 & 0.995 & 0.900 & 0.808
\\
\hline

\textbf{0.30} & 0.817 & 0.985 & 0.889 & 0.809
\\
\hline

\textbf{0.50} & 0.776 & 0.929 & 0.842 & 0.821
\\
\hline

\textbf{0.70} & 0.730 & 0.884 & 0.795 & 0.838
\\
\hline

\end{tabular*}
\end{table}

 \section{Overall Findings}
\label{sec:findings}

The central finding of this work is that \textbf{englacial layer reconstruction}
is better treated as a spatial problem than as an image-brightness one. A layer
is a geometric object: a smooth curve that runs across the radargram at a slowly
changing depth. When a detector breaks it into pieces, the brightness of each
piece is unreliable but its slope is not, staying consistent across a gap even
where the signal fades. TraceLink builds on this, predicting where a layer
continues from its current slope and linking the fragment that falls on that
trajectory; the high recall on synthetic and real radargrams
(Figures~\ref{fig:tl-syn-bars} and~\ref{fig:tl-real-bars}) shows that this
geometric continuity holds in practice. The same reasoning runs through every
stage: SyntheIce builds its radargrams from geometric priors, a shared
deformation field, a non-crossing rule, and a terrain tilt, that make the data
realistic rather than a stack of flat lines (Table~\ref{tab:dataset-comparison});
the linker weighs both depth proximity and predicted path; and XCheck requires
layers to line up across tile boundaries. Because the linking thresholds are
derived from each radargram's own spacing and gap statistics, the method adapts
to how densely the layers are packed rather than assuming a fixed geometry.
 
Recall is high everywhere because a layer rarely fades so completely that its
trajectory cannot be followed. The cost appears in precision and overlap, and
grows as the layers become denser, since closely stacked layers give the linker
more chances to attach a fragment to the wrong neighbor. This is why the
remaining errors are almost always extra splits rather than missed layers, and
why a global line fit through all of a layer's fragments in the merge pass can
rejoin the pieces the local rule left apart: the failure mode is spatial, and a
larger-scale spatial model repairs it. The threshold sweeps in
Tables~\ref{tab:iou-sensitivity} and~\ref{tab:iou-sensitivity-real} confirm this
robustness: as the IoU threshold tightens from 0.20 to 0.70, precision and F1
fall only gradually while the mean IoU of the matched layers rises rather than
drops, toward 0.96--1.00 on synthetic data and from 0.808 to 0.838 on real data.
Because the layers that survive strict matching are high-overlap traces, the
method's accuracy comes from well-placed reconstructions, not a lenient
threshold.
 
Beyond ice, the finding generalizes: when linear features in spatial data are
broken into fragments but keep a consistent local geometry, linking them by
their predicted trajectory is more reliable than by appearance. The same idea
carries over to other layered or curvilinear structures in geoscience and remote
sensing, such as sediment horizons and subsurface reflectors, where intensity is
noisy but the shape is regular. The controlled generator is a contribution in
the same spirit: it gives the spatial-data community a benchmark in which the
true geometry of every layer is known, so a method can be measured against
actual spatial structure rather than a second, imperfect annotation.

 \section{Conclusion and Future Work}
\label{sec:conclusion&futurework}

We presented an end-to-end framework for reconstructing englacial layers from
fragmented radar detections. We first found that current image generation models
cannot annotate radargrams reliably: they hallucinate layers and provide no
ground truth to validate against, because annotating englacial layers is a
spatial task rather than an image-generation one. This motivated SyntheIce, which
generates synthetic radargrams with exact per-layer ground truth, removing the
label noise that limits evaluation on real data. TraceLink then links broken
fragments into full-width layers with a slope-aware, self-tuning greedy rule
whose parameters are derived from each radargram, and a strict merge pass repairs
over-splitting. For wide radargrams, XCheck verifies annotation consistency
across adjacent tiles before reconstruction.
 
On synthetic data, TraceLink recovers layers with recall above 0.995 and F1 up
to 0.981, and it transfers to real Greenland radargrams with recall 0.93, F1
0.84, and mean IoU 0.82 at an IoU matching threshold of 0.50; The result of ranges of threshold in (Tables~\ref{tab:iou-sensitivity} and~\ref{tab:iou-sensitivity-real}) confirms these results are robust rather than dependent on a lenient threshold. To our knowledge, this is the first method to reconstruct full-width englacial layers from fragments using slope, validated on both controlled synthetic data and real expert annotations from the Greenland ice sheet. Because the dominant error is extra splits rather than missed layers, a short manual clean-up closes the remaining gap at far lower cost than re-tracing from scratch; the main limitation is that the synthetic terrain tilt is mild, up to about $3^{\circ}$, compared with real Greenland layers, so steeply dipping settings are under-tested.

In future work, we plan to extend SyntheIce to richer tilt across the whole
profile, move toward a learned non-parametric linker, trace consistent layers
across multiple consecutive real radargrams, and apply TraceLink to annotations
from semi-automated detectors. We also plan to generate raw radargrams from the
synthetic masks using machine-learning techniques, then train a model on this
paired synthetic data and apply it to real radargrams to produce annotations. At
scale, such annotations would support the study of past climate and the
ice-melt and flow history of ice sheets and glaciers, and help predict their
future behaviour that contributes to overall climate of the planet.

\section*{Acknowledgment}
This work has been funded by the NSF HDR Institute for Harnessing Data and Model Revolution in the Polar Regions (iHARP), Award \#2118285. We acknowledge Dr. Nicholas D. Holschuh for providing ice layer expert-labeled dataset.

\bibliographystyle{IEEEtran}
\bibliography{sample-base}

\end{document}